\documentclass[10pt,twocolumn,letterpaper]{article}

\usepackage{cvpr}              

\definecolor{cvprblue}{rgb}{0.21,0.49,0.74}
\usepackage[pagebackref,breaklinks,colorlinks,allcolors=cvprblue]{hyperref}
\usepackage{booktabs}
\usepackage{multirow}
\usepackage{siunitx}
\usepackage{adjustbox}
\usepackage{algorithm}
\usepackage{algorithmic}
\usepackage{caption}
\usepackage{amsmath}
\usepackage[accsupp]{axessibility}  

\def\paperID{***} 
\def\confName{CVPR}
\def\confYear{2026}

\title{MagicPrompt: Ultra-Lightweight Prompt Tuning
for Video Generation}

\author{
Yinhan Zhang\textsuperscript{1,3*}, \quad 
Dingwei Tan\textsuperscript{2*}, \quad 
Xianghao Kong\textsuperscript{1}, \quad 
Yue Ma\textsuperscript{1,$\ddagger$}, \quad 
Yeying Jin\textsuperscript{3\dag}, \quad 
Anyi Rao\textsuperscript{1\dag}
\\
\\
\textsuperscript{1} HKUST,
\textsuperscript{2} HKUST(GZ),
\textsuperscript{3} Tencent
\\
\\
Project Page: \url{https://github.com/YinHan-Zhang/MagicPrompt}
}

\begin{document}

\maketitle

\newcommand\blfootnote[1]{
  \begingroup
  \renewcommand\thefootnote{}\footnote{#1}
  \addtocounter{footnote}{-1}
  \endgroup
}

\blfootnote{$*$ Equal contribution.}
\blfootnote{$\ddagger$ Project Leader.}
\blfootnote{$\dag$ Corresponding author.}
\blfootnote{This work originated from an internship at Tencent.}

\begin{abstract}

Large-scale video diffusion models (VDMs) deliver strong generation performance, but full fine-tuning for downstream tasks incurs prohibitive computational costs. Existing parameter-efficient fine-tuning (PEFT) methods have two critical flaws on billion-scale models: they still require substantial trainable parameters, and reward-based training suffers from noise-induced optimization instability in condition-guided tasks.
We propose MagicPrompt, a lightweight framework that achieves extreme parameter efficiency and stable reward optimization. It first adopts Attention-Embedded Prompt Tuning, which steers generation via lightweight soft prompts with orders of magnitude fewer parameters while preserving pre-trained knowledge. It further introduces Dual-Space Reward Feedback Optimization, which uses self-supervised latent objectives to improve condition-guided reward training. Experiments show MagicPrompt reaches competitive performance with less than 1\% trainable parameters and notably reduces training costs.

\end{abstract}    

\section{Introduction}
The advent of large-scale video diffusion models (VDMs)~\cite{liu2025survey,zheng2026forecast,ma2026group,ma2024followpose,ma2025followcreation,ma2026fastvmt,ma2025followyourmotion} has marked a paradigm shift in generative artificial intelligence, demonstrating remarkable capabilities in synthesizing high-quality, temporally coherent videos from textual prompts. However, the sheer scale of these models presents a significant barrier to practical adoption, as adapting a pre-trained VDM to downstream tasks typically requires full fine-tuning that updates billions of parameters. This approach is not only computationally prohibitive due to immense memory and storage demands but also risks destabilizing the model's original generative distribution. To mitigate these costs, \textbf{Parameter-Efficient Fine-Tuning (PEFT)} has emerged as a critical direction, aiming to customize massive models by updating only a minimal subset of parameters while freezing the pre-trained backbone. Despite this promise, adapting video diffusion models remains an open challenge characterized by interconnected bottlenecks: existing methods often suffer from residual parameter inefficiency and computational overhead, while still demanding large-scale datasets that are scarce for video tasks. More critically, under limited-data settings, these approaches are prone to overfitting and catastrophic forgetting, where task-specific updates degrade the model's original generative diversity. Addressing these challenges requires a delicate balance between adaptation flexibility and structural stability.

\begin{figure*}[htbp]
    \centering
    \includegraphics[width=\linewidth, keepaspectratio]{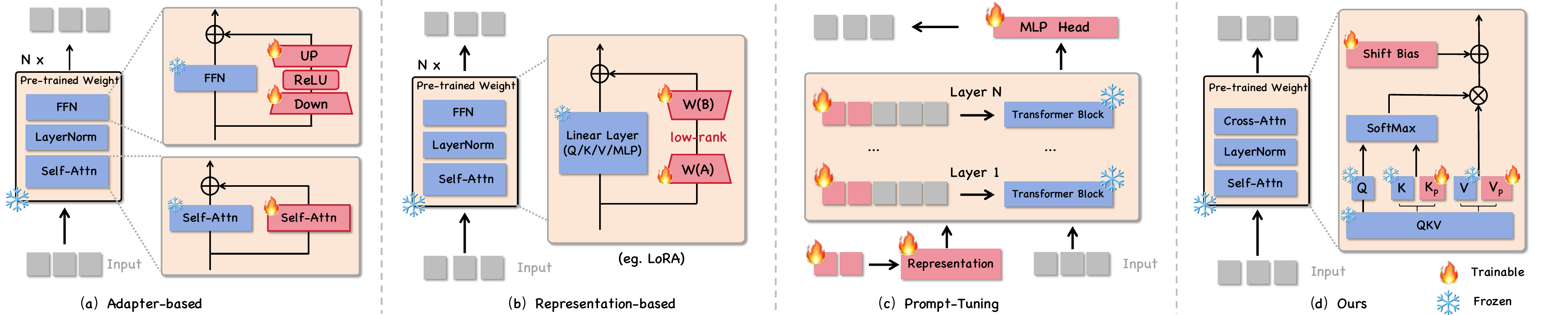}
    \caption{\textbf{Comparison of PEFT paradigms.} 
    (a) Adapter-based methods insert auxiliary modules in parallel alongside the backbone. 
    (b) Representation-based methods (e.g., LoRA) approximate weight updates via low-rank decomposition.
    (c) Prompt-Tuning methods add trainable soft prompts to each layer and combine with the layer input.
    (d) Our method employs attention-embedded prompt tuning, injecting trainable prompts directly into attention mechanisms. 
    }
    \label{fig: motivation}
\end{figure*} 
As illustrated in Fig.~\ref{fig: motivation}, current strategies for efficient adaptation generally fall into three paradigms, drawing inspiration from Vision Transformers (ViT) and Diffusion models. 
\textbf{(a) Adapter-based Methods} insert lightweight modules into the pre-trained backbone. Building on ViT adaptations~\cite{ST-Adapter, AdaptFormer, jie2024convolutional}, diffusion counterparts such as ControlNet~\cite{controlnet, ipadapter} and video-specific works like SimDA~\cite{simda} introduce auxiliary networks for control signals or temporal features. However, these auxiliary modules introduce millions of trainable parameters, incurring substantial memory and computational overhead on large-scale models. 
\textbf{(b) Representation-based Methods} modify weight representations~\cite{lian2022scaling, jie2023fact} directly rather than adding modules. LoRA~\cite{hu2021lora} employs low-rank decomposition, while RepAdapter~\cite{RepAdapter} enables re-parameterization for inference efficiency. Yet, such \textbf{intrusive weight modifications} can disrupt pre-trained knowledge, leading to artifacts like inter-frame jitter. 
\textbf{(c) Prompt-Tuning Methods}~\cite{vpt, apt, PATT, DAPE, FedPerfix, sohn2023visual} guide attention with learnable prompts while freezing backbones, achieving high parameter efficiency in ViTs. However, their application to video diffusion models remains significantly under-explored, as the iterative denoising process poses unique challenges for maintaining temporal consistency and optimization stability.
Unlike prior works, \textbf{(d) our approach} steers generation without modifying model weights, achieving comparable performance with parameter efficiency.

Despite this diversity, two critical limitations persist that hinder practical deployment.
\textit{(1) Huge model trainable parameters and high resource demands.} 
Existing PEFT methods still require training millions of parameters, which becomes prohibitive for models with more than 10 billion parameters. 
More critically, when scaling to massive video diffusion architectures, the cumulative memory footprint and computational overhead of these auxiliary modules remain substantial, rendering them impractical for practitioners with resource-constrained hardware. 
\textit{(2) High Noise Incompatibility in condition-guided reward optimization.} 
While recent works like GRPOs~\cite{grpo, DanceGRPO} apply reinforcement learning for alignment, they typically require full-model updates and group sampling, which are prohibitive for video diffusion. It also requires large labeled datasets and separate reward model training, which contradicts the goal of parameter and data efficiency. 
Furthermore, applying pixel-space rewards directly to latent representations is incompatible with high-noise denoising stages, causing optimization instability and a lack of timestep insensitivity.
These gaps underscore the need for a more efficient framework that combines extreme parameter and data efficiency with stable optimization.
Specifically, there is a pressing need for an \textit{attention-embedded prompt tuning} mechanism that steers generation with minimal parameters to enhance attention performance without modifying pre-trained knowledge, alongside a \textit{dual-space reward feedback optimization}, training via self-supervised latent objectives that enhance condition reward optimization.
Addressing these challenges motivates our proposed MagicPrompt, which seeks to balance adaptation flexibility with structural efficiency for high-quality video generation.

To bridge these gaps, we present MagicPrompt, a novel prompt-tuning framework engineered to achieve superior parameter efficiency and generalization capability for video diffusion models. 
Our core motivation is that strategic placement of lightweight soft prompts can effectively steer the model's attention mechanisms without modifying its core weights. 
This paradigm enables efficient adaptation by leveraging pre-trained knowledge for downstream tasks while helping mitigate the risk of catastrophic forgetting.

Architecturally, MagicPrompt operationalizes two design principles to systematically address the identified challenges: non-intrusive adaptation to preserve efficiency, and reward-guided optimization to promote alignment. 
To resolve parameter inefficiency and high resource demands, we eschew direct weight modification in favor of \textit{Attention-Embedded Prompt Tuning}, injecting trainable soft prompts directly into attention mechanisms where feature integration occurs. 
In attention layers, learnable prompts prepended to Key and Value projections enable domain adaptation via attention redistribution, while frozen Query projections and backbone weights safeguard original generative capabilities.
This restriction of updates to lightweight prompts reduces the number of trainable parameters by orders of magnitude compared to LoRA, enabling billion-scale adaptation on consumer hardware while inherently preventing forgetting. 
However, condition-video alignment requires specific reward models and is unstable during high-noise stage optimization.
Thus, we integrate a \textit{Dual-Space Reward Feedback Optimization} that supervises training via pixel-space perceptual rewards (HPS/MPS) for visual quality and latent-space objectives that enforce efficient denoising and significantly enhance generalization in condition-guided generation scenarios. 
Empirical validation confirms that MagicPrompt achieves competitive performance on video generation tasks while utilizing significantly fewer trainable parameters than LoRA and adapter-based methods. 
The contributions of this work are threefold:
\begin{itemize}
    \item We propose \textbf{MagicPrompt}, a lightweight parameter-efficient tuning framework tailored for video diffusion models, enabling billion-scale model customization on resource-constrained hardware.
    
    \item We introduce \textit{Attention-Embedded Prompt Tuning}, which injects trainable visual and textual prompts directly into the attention layers of video diffusion models. To overcome high noise incompatibility in reward training on the condition-guided generation task, we introduce a \textit{Dual-Space Reward Feedback Optimization}. This non-intrusive approach preserves performance with orders of magnitude fewer parameters than existing methods and improves condition-guided reward training.
    
    \item Extensive experiments across text-to-video, image-to-video, and control-to-video tasks demonstrate that MagicPrompt achieves competitive performance with strong scalability across different model sizes.
\end{itemize}

\section{Related Work}

\subsection{Video Diffusion Models}
Diffusion models have revolutionized generative tasks by iteratively denoising random noise into structured data~\cite{ddpm, ddim}. 
Formally, the forward process diffuses data $x_0$ into noise $x_T$ via a Markov chain, while the reverse process learns to recover $x_0$ by minimizing the noise prediction error $\mathcal{L} = \mathbb{E}_{x_0, t, \epsilon} [ \| \epsilon - \epsilon_\theta(x_t, t) \|^2 ]$. 
Extending this paradigm to video generation introduces significant complexity, as models must capture both spatial fidelity and temporal coherence within spatio-temporal tensors $V \in \mathbb{R}^{T \times H \times W \times C}$.
Video diffusion models have evolved substantially since their inception, progressing from early 3D U-Net-based spatio-temporal modeling to the now-prevalent Transformer-centric architectural paradigm.
Alongside architectural advances, the field has progressed in two complementary directions: reducing the computational cost of video synthesis and enhancing controllability for flexible conditional generation.
Pioneering works established the 3D U-Net framework for video diffusion~\cite{ho2022video, wan, ma2025controllable,ma2025followfaster,ma2024followyouremoji,song2026vista,song2026streamingeffect,gao2026pai,song2024processpainter, Deng_2026_CVPR, zhang2026instanceanimator, zhang2025magiccolor, qiu2024tfb, qiu2025duet, qiu2025DBLoss, zhang2026tea, zhang2025magiccolor}, and the scalable DiT paradigm later drove the shift to Transformer backbones, adopted by subsequent models such as Latte and CogVideo~\cite{peebles2023scalableDiT, Latte, cogvideo}.
On the efficiency front, VideoLDM and VideoFusion proposed latent space optimization and multi-scale diffusion strategies, respectively~\cite{videoldm, luo2023videofusion}; for controllable generation, Make-A-Video, AnimateDiff, and ControlVideo have successively advanced text-guided synthesis and explicit motion control~\cite{singer2022make, deng2024compact, chen2025contextflow, wang2024taming, feng2025dit4edit, wang2024cove, ma2025magicstick, AnimateDiff, ControlVideo}.
However, as these models scale to billions of parameters to achieve higher fidelity, full fine-tuning becomes increasingly prohibitive, necessitating efficient adaptation strategies.

\subsection{PEFT for Generative Models}
Parameter-Efficient Fine-Tuning (PEFT) has become indispensable for adapting large generative models, addressing the prohibitive computational costs and storage requirements of full fine-tuning.
In the realm of image generation, seminal works like \textit{DreamBooth}~\cite{DreamBooth} enable personalized synthesis by fine-tuning specific embeddings, while \textit{DiffLoRA}~\cite{DiffLoRA} adapts low-rank decomposition for diffusion pipelines.
These methods typically fall into three categories: \textbf{Representation-based} approaches, such as \textit{LoRA}~\cite{hu2021lora}, approximate weight updates via low-rank matrices using <1\% trainable parameters; \textbf{Adapter-based} methods~\cite{houlsby2019parameter} insert lightweight modules between layers for task-specific adaptation; and \textbf{Prompt-based} techniques like \textit{Prefix Tuning}~\cite{prefixtuning} learn continuous embeddings to steer generation without modifying model weights.
Extending these paradigms to \textbf{video diffusion models} introduces unique challenges due to temporal dynamics and increased computational load.
Recent adaptations include video-specific LoRA variants and adapter-based frameworks such as \textit{SimDA}~\cite{simda} and \textit{VACE}~\cite{vace}, which incorporate modules to handle spatio-temporal features.
While these methods reduce training costs by orders of magnitude and preserve pre-trained capabilities, they remain limited in few-shot scenarios.
Specifically, existing video PEFT methods often struggle with the computational resource demands of generative diversity and text-video alignment when labeled data is scarce.
This gap highlights the need for a more robust tuning framework that balances parameter efficiency with strong generalization capabilities in data-scarce regimes.

\begin{figure*}[htbp]
\centering
\includegraphics[width=\linewidth, keepaspectratio]{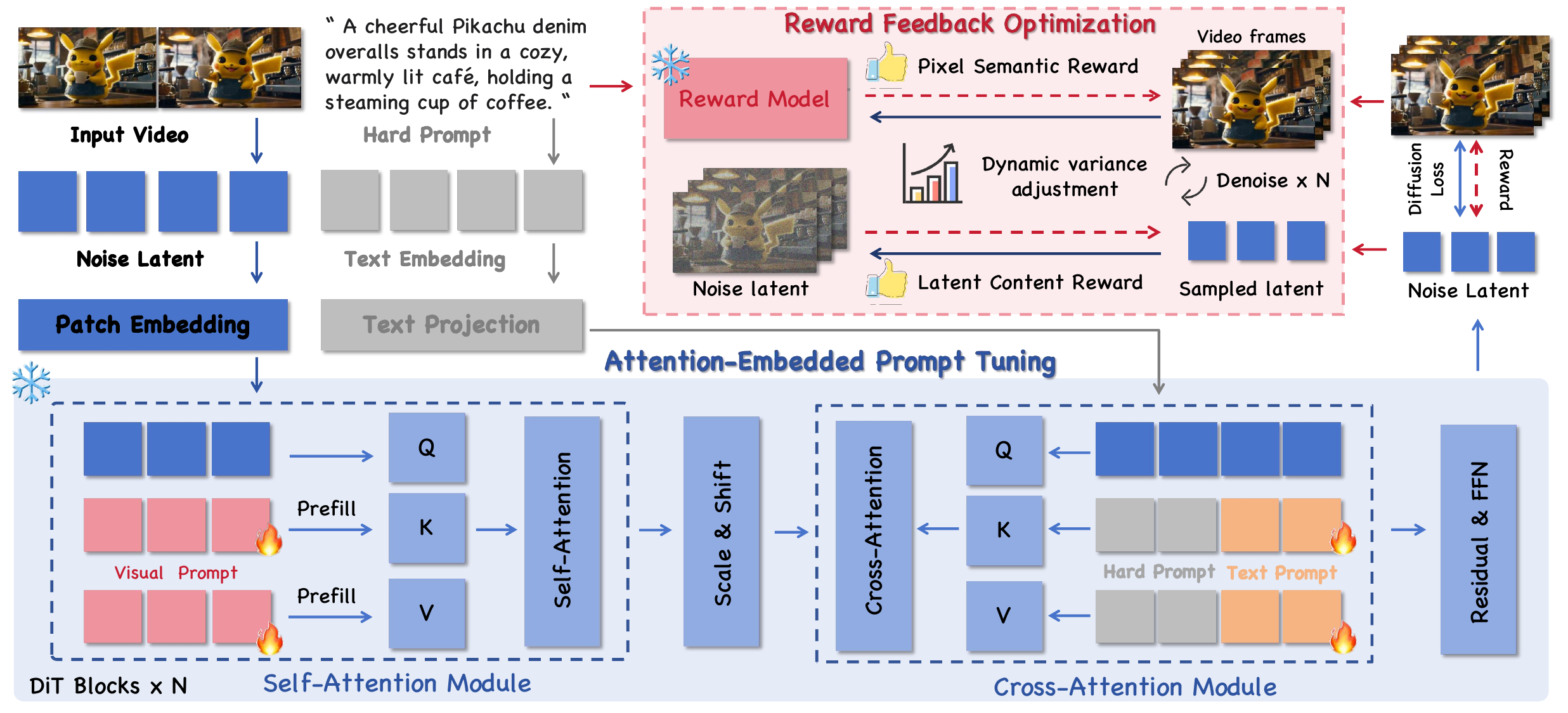}
\caption{\textbf{Framework overview of MagicPrompt.} Visual soft prompt prefill with latent feature in the key and value in the self-attention module, and textual soft prompt combined with text embedding in the cross-attention module to reweight the attention distribution. To better align with human preference, the reward feedback optimization is guided by both pixel-space semantic reward and latent objectives.
}
\label{fig: framework}
\end{figure*} 

\section{Method}

Our method consists of two core components, as illustrated in Figure~\ref{fig: framework}. 
In Section~\ref{sec: problem}, we describe the definition of the problem.
Section~\ref{sec: prompt_tuning} introduces the Attention-Embedded Prompt Tuning method, which strategically inserts trainable soft prompts into both self-attention and cross-attention modules to efficiently adapt the pre-trained model to downstream tasks with minimal parameter overhead. 
Section~\ref{sec: reward_training} presents the Dual-Space Reward Feedback Optimization, which provides pixel-space rewards via HPS and MPS scores, and latent-space rewards through diffusion latent analysis to guide soft prompt learning toward high-quality generation.

\subsection{Problem Definition}
\label{sec: problem}
Let $\mathcal{M}_\theta$ denote a pre-trained video diffusion model with \textit{frozen} parameters $\theta$, generating video frames $V$ conditioned on input $C$. Given target data $\mathcal{D}_{\text{target}}$, our goal is to adapt $\mathcal{M}_\theta$ to the target domain by learning \textit{only} task-specific soft prompts $\mathcal{P} = \{P_{txt}, P_{vis} , P_{bias}\}$. $P_{txt} \in \mathbb{R}^{L \times d}$ (textual soft prompts) are injected before text embeddings to guide cross-attention, and $P_{vis} \in \mathbb{R}^{L \times d}$ (visual soft prompts) are inserted into self-attention key-value sequences with shift biases term.  
Crucially, $\theta$ remains \textit{untrained} while $\mathcal{P}$ is optimized to minimize:  
\begin{equation}  
\mathcal{L} = \mathbb{E}_{(V_0, C) \sim \mathcal{D}_{\text{target}}} \left[ \| \epsilon - \epsilon_{\theta, \mathcal{P}}(V_t, t, C) \|^2 \right] ,
\end{equation}  
where $\epsilon_{\theta, \mathcal{P}}$ denotes the noise prediction with prompt-enhanced attention and $V_t$ is the noisy latent/video representation obtained from $V_0$
at timestep $t$, and $\epsilon$ denotes the sampled Gaussian noise.

\subsection{Attention-Embedded Prompt Tuning}
\label{sec: prompt_tuning}

\subsubsection{Visual Prompt}  
In video diffusion transformers, self-attention processes spatiotemporal tokens, where long-range temporal dependencies are essential for motion coherence. However, directly tuning backbone parameters incurs prohibitive computational costs while risking catastrophic forgetting of pre-trained dynamics. 
To resolve this, we inject task-specific visual prompts $P_{vis}^{\text{key}}, P_{vis}^{\text{value}} \in \mathbb{R}^{L_{vis} \times d}$ into the key-value sequences of each DiT block:

\begin{equation}
    \begin{aligned}
        \tilde{\mathbf{K}}_{vis} &= [\mathbf{K} \oplus P_{vis}^{\text{key}}], \mathbf{K} \in \mathbb{R}^{B\times Seq \times d} , P_{vis}^{\text{key}} \in \mathbb{R}^{L_{vis} \times d}, \\
        \tilde{\mathbf{V}}_{vis} &= [\mathbf{V} \oplus P_{vis}^{\text{value}}], \mathbf{V} \in \mathbb{R}^{B\times Seq \times d} , P_{vis}^{\text{value}} \in \mathbb{R}^{L_{vis} \times d},
    \end{aligned}
\end{equation}
where $\oplus$ denotes concatenation, $\tilde{\mathbf{K}}_{vis}$ and $\tilde{\mathbf{V}}_{vis}$ in attention layer apply to each DiT Block.
This design preserves pre-trained spatial features while enabling dynamic attention redistribution across frames. 

\begin{equation}
     \mathbf{X} = \text{Self-Attention}(\mathbf{Q}, \tilde{\mathbf{K}_{vis}} ,\tilde{\mathbf{V}_{vis}})
\end{equation}
Crucially, the visual prompts serve as temporal anchors: during denoising, queries from motion-critical regions (e.g., moving objects) develop stronger attention to $P_{vis}$, thereby explicitly guiding the model to prioritize relevant temporal patterns without modifying the frozen backbone. Consequently, the model achieves efficient adaptation with minimal parameter overhead ($L \cdot d \ll |\theta|$), directly addressing the resource constraints of billion-parameter video generators.  

\subsubsection{Shift Bias}
While visual prompts redistribute attention, they introduce latent space distribution shifts that manifest as frame inconsistency (e.g., jittery motions or abrupt scene transitions). This occurs because the concatenated prompts alter the statistical properties of the attention output, destabilizing the denoising trajectory across frames. 
To mitigate this, we introduce a learnable shift bias mechanism that calibrates the latent distribution through 
\begin{equation}
    \mathbf{X}' = \gamma \cdot \mathbf{X}  + \beta, \quad \gamma \in \mathbb{R}^{1 \times 1 \times d} ,\beta \in \mathbb{R}^{1 \times 1 \times d}
\end{equation}
where $\gamma$ and $\beta$ are the layer-specific scale and bias trainable vectors. The scale $\gamma$ amplifies weak but meaningful attention patterns (e.g., subtle object trajectories) that would otherwise be suppressed by noise, while the bias $\beta$ corrects systematic offsets in denoising trajectories (e.g., consistent motion direction errors). This dual calibration ensures that model weights remain bounded, thereby helping improve temporal coherence. Thus, the shift bias alleviates frame inconsistency introduced by visual prompts, forming a closed-loop optimization for stable video generation.

\subsubsection{Textual Prompt}
In cross-attention, text embeddings $\mathbf{Context} \in \mathbb{R}^{L_c \times d}$ condition the diffusion process by aligning latent tokens with language semantics. However, unmodified embeddings often fail to precisely steer attention toward semantically critical regions, resulting in misaligned generations (e.g., the prompt "a dog running in a park" producing static dogs). To address this, we prepend task-aware textual prompts $P_{txt} \in \mathbb{R}^{L_{txt} \times d}$ to the text embeddings, forming $\tilde{\mathbf{Context}} = [P_{txt} \oplus \mathbf{Context}]$, where $L_{txt} \ll L_c$ denotes the prompt length. The augmented embeddings are then projected into key-value sequences for cross-attention:

\begin{equation}
    \begin{aligned}
        \tilde{\mathbf{K}}_{txt} &= \tilde{\mathbf{Context}}\mathbf{W}_K, \\
        \tilde{\mathbf{V}}_{txt} &= \tilde{\mathbf{Context}}\mathbf{W}_V,
    \end{aligned}
\end{equation}
where $\mathbf{W}_K, \mathbf{W}_V \in \mathbb{R}^{d \times d}$ are frozen key/value projection matrices in the cross-attention layers of the pre-trained diffusion model. The cross-attention output is computed as:

\begin{equation}
    \mathbf{X}_{cross} = \text{Softmax}\left(\frac{\mathbf{Q} \tilde{\mathbf{K}}_{txt}^\top}{\sqrt{d}}\right) \tilde{\mathbf{V}}_{txt}
\end{equation}

During optimization, $P_{txt}$ learns to function as a semantic filter: it suppresses attention to irrelevant tokens while amplifying attention to relevant tokens. This reshapes the attention distribution such that queries from motion-critical latent regions develop a stronger affinity toward contextually salient text tokens. Crucially, by prepending prompts rather than modifying the original embeddings, we preserve the integrity of pre-trained language representations while enabling task-specific adaptation. This design achieves robust text-video alignment with minimal parameter overhead ($L_{txt} \cdot d \ll |\theta|$), directly addressing semantic misalignment without altering the frozen text encoder.

\subsection{Dual Reward Feedback Optimization}
\label{sec: reward_training}

\subsubsection{Pixel-Space Reward} 
To directly optimize perceptual quality and semantic alignment, we compute rewards in pixel space by decoding denoised latents to video frames. Following the practice of DRaFT~\cite{daft, alignt2v}, we first sample key frames from the denoised video at low-noise timesteps ($t \geq 3T/4$, in inference mode), where visual semantics are sufficiently clear for reliable evaluation. These frames are then decoded via the VAE to obtain RGB frames $\hat{V} \in \mathbb{R}^{F \times H \times W \times 3}$. We employ two established perceptual metrics: the Human Preference Score (HPS) measures text-image alignment by evaluating CLIP similarity between generated frames and the input text prompt, while the Motion Preference Score (MPS) assesses temporal coherence by computing optical flow consistency across consecutive frames. The pixel-space reward is defined as follows:
\begin{equation}
    R_{\text{pixel}} = \alpha \cdot \text{HPS}(\hat{V}, C) + (1-\alpha) \cdot \text{MPS}(\hat{V}), 
\end{equation}
where $\alpha$ balances semantic fidelity and motion quality. This reward provides direct supervision on visual quality but suffers from three limitations: first, computing rewards at early diffusion steps (high-noise stages) yields unreliable signals due to semantic ambiguity; second, reward values fluctuate significantly across timesteps, destabilizing training; third, deploying external reward models (e.g., HPS) incurs additional computational overhead. These limitations motivate our latent-space reward design to combine with the pixel reward.

\begin{figure*}[htbp]
\centering
\includegraphics[width=\linewidth, keepaspectratio]{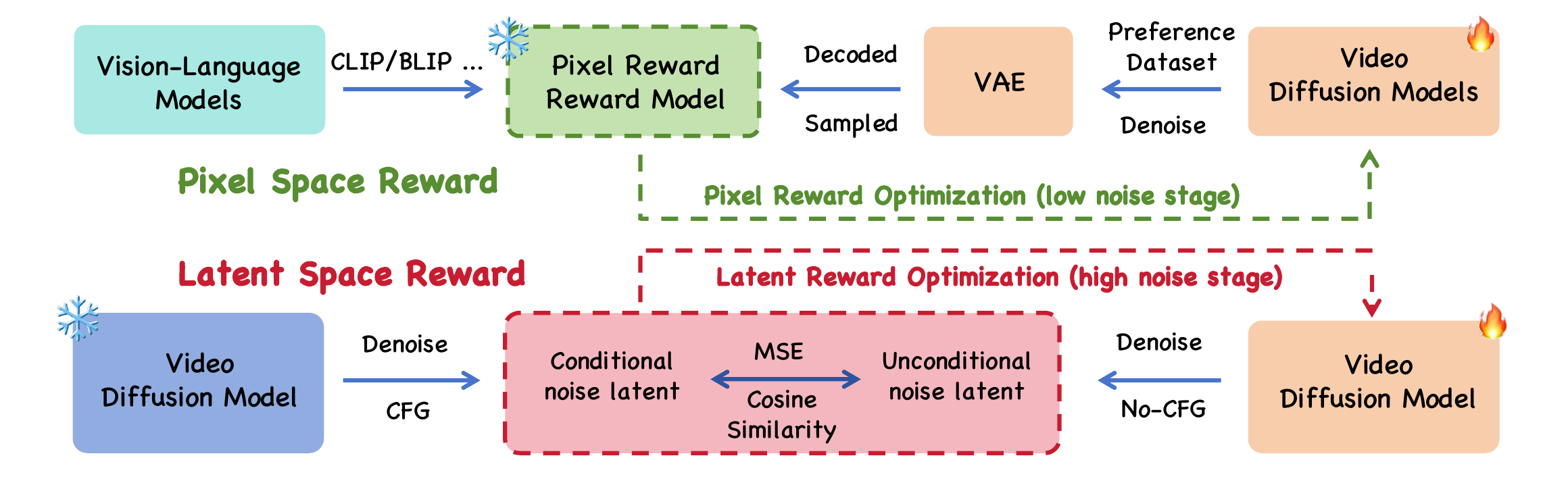}
\caption{\textbf{Overview of dual reward feedback optimization.} To enhance generalization in reward training, MagicPrompt integrates regularization signals from two domains. 
\textit{Pixel Space:} Perceptual rewards (HPS, MPS) evaluate decoded frames to ensure visual fidelity. 
\textit{Latent Space:} Complementary objectives enforce efficient denoising trajectories and robust conditioning via CFG.}
\label{fig: reward}
\end{figure*} 

\subsubsection{Latent-Space Reward} 
To overcome the pixel-space reward cannot be calculated during high-noise stages and the computational burden of pixel-space rewards, we propose a latent-space reward that leverages intrinsic properties of the diffusion process without requiring external reward models. 
Our design is grounded in two empirical observations: 
(1) Classifier-free guidance (CFG) consistently yields higher generation quality than unconditional sampling, indicating that CFG trajectories encode richer conditional semantics;
(2) Latent representations become more coherent as noise level decreases, meaning the denoising process itself implicitly encodes quality progression along the timestep dimension.
Based on these insights, we construct our latent reward by comparing two parallel denoising trajectories derived from the same input at the same timestep $t$: a CFG-guided trajectory and an unconditional trajectory. Specifically, we extract latent representations from both trajectories at timestep $t$, denoted $z_{t}^{\text{cfg}}$ and $z_{t}^{\text{uncond}}$, and compute their similarity as the reward signal:
\begin{equation}
    R_{\text{latent}} = 1 - \operatorname{MSE}\left(z_{t}^{\text{cfg}}, z_{t}^{\text{uncond}}\right),
\end{equation}
This formulation follows two core design principles. First, the CFG trajectory at timestep $t$ serves as the target quality level we aim for the model to achieve. Second, the unconditional trajectory at the same timestep acts as a baseline that represents unguided, noisier generation. Minimizing the distance between the two encourages the model to reach CFG-level quality even with weak guidance, effectively accelerating training convergence.
Crucially, this reward is robust to noise level variations and timestep-agnostic: it operates on relative similarity between latents rather than absolute pixel values, so it remains effective across all denoising stages and eliminates the need for carefully designed reward scheduling.
The final combined reward integrates both objectives:
\begin{equation}
    R = \lambda_{\text{pixel}} R_{\text{pixel}} + \lambda_{\text{latent}} R_{\text{latent}},
\end{equation}
where $\lambda_{\text{pixel}}$ and $\lambda_{\text{latent}}$ balance pixel-level perceptual fidelity and latent-space semantic consistency. This dual reward mechanism stabilizes training while ensuring high-quality video generation.
\section{Experiment}

\subsection{Experimental Setup}
\label{sec:experimental_setup}

\paragraph{Setup.} We evaluate our method on three video generation tasks. For text-to-video and image-to-video generation, we use the OpenVid dataset~\cite{OpenVid}, which contains diverse video-text pairs covering various scenes and motions. We randomly split the dataset into training and test sets with a 9:1 ratio to ensure fair evaluation. For control-to-video generation, we train on the OpenHumanVid dataset~\cite{OpenHumanVid}, which provides human-centric videos with control signals (e.g., canny, depth), and evaluate on the TikTok dataset~\cite{tiktokdataset}, a challenging benchmark featuring complex human motions and diverse backgrounds. Our pre-trained weights are from Wan2.1-Fun-InP and Wan2.2-TI2V-5B~\cite{wan}. For fair comparison, we standardize the training configuration across all methods. \textbf{LoRA} is configured with Rank=64, $\alpha$=32, and learning rates of 1e-4 (1.3B/5B) and 3e-5 (14B). \textbf{MagicPrompt} uses 64 soft prompt tokens with learning rates of 5e-3 (1.3B/5B) and 1e-3 (14B).

\paragraph{Evaluation Metrics.} 
We employ a comprehensive set of four metrics to assess generation quality from multiple perspectives. 
For semantic alignment, we compute the CLIP Score~\cite{clip} between generated frames and input conditions (text prompts or control signals), where higher values indicate better text-video or control-video correspondence. For perceptual quality, we adopt LPIPS~\cite{lpips} to measure the perceptual similarity between generated and reference frames, with lower scores reflecting higher visual fidelity. 
For distribution-level realism, we use Fréchet Inception Distance (FID)~\cite{fid} to compare feature distributions of generated and real videos, where lower values denote better image quality. 
Finally, for temporal coherence, we report Fréchet Video Distance (FVD)~\cite{fvd} to assess motion smoothness and video-level consistency, with lower scores indicating superior temporal dynamics. 
All metrics are computed on held-out test sets, with higher CLIP Score and lower LPIPS/FID/FVD indicating better overall performance.

\begin{figure*}[htbp]
\centering
\includegraphics[width=\linewidth, keepaspectratio]{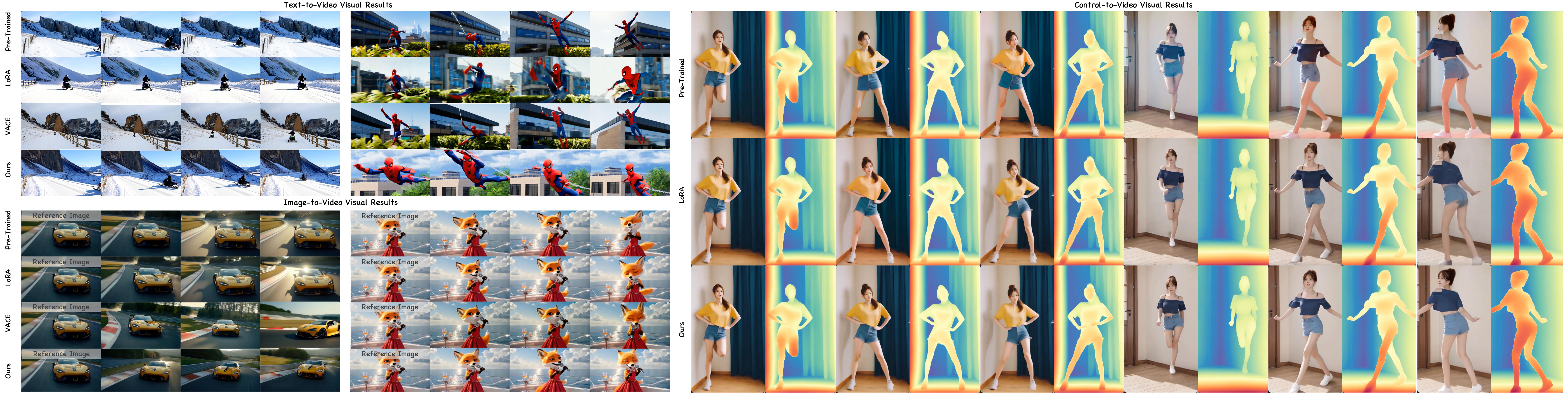}
\caption{\textbf{Qualitative comparison with baselines}. Our method demonstrates superior temporal coherence and visual fidelity across T2V, I2V, and Control2V tasks. Unlike baselines exhibiting motion blur and structural distortion, our method preserves fine details and ensures consistent motion dynamics.}
\label{fig: visual_comparison}
\end{figure*} 

\subsection{Quantitative Experiments}
We conduct quantitative evaluations across T2V, I2V, and Control2V test sets, comparing MagicPrompt against three representative baselines. 
As shown in Table~\ref{tab: comparison}, MagicPrompt achieves competitive or superior performance across all tasks while updating the fewest parameters. 

\begin{table*}[tb]
  \centering
  \resizebox{\textwidth}{!}{
  \begin{tabular}{@{}lcccccccccccc@{}}
    \toprule
    \multirow{2}{*}{Method} & 
    \multicolumn{4}{c}{T2V} & 
    \multicolumn{4}{c}{I2V} & 
    \multicolumn{4}{c}{Control2V} \\
    \cmidrule(lr){2-5} \cmidrule(lr){6-9} \cmidrule(lr){10-13}
    & CLIP $\uparrow$ & LPIPS $\downarrow$ & FID $\downarrow$ & FVD $\downarrow$
    & CLIP $\uparrow$ & LPIPS $\downarrow$ & FID $\downarrow$ & FVD $\downarrow$
    & CLIP $\uparrow$ & LPIPS $\downarrow$ & FID $\downarrow$ & FVD $\downarrow$ \\
    \midrule
    Pre-Trained & 0.66 & 0.69 & 103.53 & 824.53 & 0.89 & 0.38 & 57.72 & 299.81 & 0.85 & 0.40 & 30.35 & 488.72 \\
    LoRA & 0.68 & \textbf{0.66} & 101.76 & 853.56 & 0.91 & \textbf{0.35} & 43.19 & \textbf{269.02} & 0.86 & 0.39 & \textbf{26.09} & 419.82 \\
    VACE & 0.65 & 0.71  & 99.64 & 760.33 & 0.88 & 0.44 & 43.04 & 354.67 & 0.86 & \textbf{0.41} & 39.29 & 464.45  \\
    \textbf{Ours} & \textbf{0.68} & 0.66 & \textbf{86.56} & \textbf{637.46} & \textbf{0.92} & 0.37 & \textbf{29.23} & 274.88 & \textbf{0.87} & 0.39 & 34.43 & \textbf{412.82} \\
    \bottomrule
  \end{tabular}%
  }
  \vspace{0.5em}
  \caption{Quantitative comparison with baseline methods across three tasks: Text-to-Video (T2V), Image-to-Video (I2V), and Control-to-Video (Control2V). T2V and I2V use the 1.3B model, whereas Control2V adopts 14B model.}
  \label{tab: comparison}
\end{table*}

\paragraph{Analysis of Comparison.}
As shown in Tab.~\ref{tab: comparison}, in Text-to-Video generation, our method matches LoRA in CLIP Score and LPIPS, while substantially outperforming all baselines in distribution quality, achieving the best FID and FVD. 
This indicates that our attention-embedded prompts effectively adapt semantic alignment without compromising visual fidelity. 
In Image-to-Video, MagicPrompt attains the highest CLIP Score and the lowest FID, demonstrating strong preservation of input image semantics. While LoRA shows a marginal advantage in FVD, our method remains highly competitive with significantly fewer trainable parameters. 
For Control-to-Video, MagicPrompt achieves the best FVD, outperforming VACE and LoRA while maintaining comparable CLIP and LPIPS scores, highlighting its robustness in handling complex control signals.
Critically, these results are achieved with orders of magnitude fewer trainable parameters than LoRA or VACE. This efficiency-performance trade-off makes MagicPrompt particularly suitable for resource-constrained adaptation of billion-scale video diffusion models.

\begin{table}[tb]
  \centering
  \footnotesize 
  \setlength{\tabcolsep}{5pt}
  \resizebox{\linewidth}{!}{
  \begin{tabular}{@{}lcccccc@{}}
    \toprule
    Method & CLIP $\uparrow$ & LPIPS $\downarrow$ & SSIM $\uparrow$ & PSNR $\uparrow$ & FID $\downarrow$ & FVD $\downarrow$ \\
    \midrule
    Latent Reward   & 0.92 & 0.27 & \textbf{0.81} & 19.57 & 21.46 & 217.91 \\
    Pixel Reward    & 0.89 & 0.29 & 0.77 & 18.90 & 18.81 & 142.06 \\
    Dual Reward     & \textbf{0.93} & \textbf{0.27} & 0.79 & \textbf{19.64} & \textbf{17.92} & \textbf{141.44} \\
    No Reward       & 0.92 & 0.37 & 0.71 & 18.21 & 29.12 & 274.88 \\
    \bottomrule
  \end{tabular}}
  \vspace{0.5em}
  \caption{Comparison of different reward strategies and training configurations. Our full method achieves competitive performance compared to Pixel Reward and Latent Reward. $\uparrow$ indicates higher is better, $\downarrow$ indicates lower is better.}
  \label{tab: reward_comparison}
\end{table}

\paragraph{Analysis of Reward Feedback.}
To isolate the impact of reward strategies, we maintain the trainable parameters fixed as visual and textual soft prompts throughout this experiment.
As detailed in Table~\ref{tab: reward_comparison}, our Dual-Space Reward Feedback Optimization enables the model to maintain robust condition-video alignment.
Our ablation study demonstrates that pixel-space and latent-space rewards each possess distinct and complementary strengths. When integrated into the Dual Reward framework, their combined effect achieves the best CLIP, PSNR, FID, and FVD, while remaining competitive in LPIPS and SSIM. 
This demonstrates that latent-space objectives effectively regularize the early denoising trajectory, allowing for fewer sampling steps without collapsing generation quality.
Furthermore, compared to No Reward (without reward training), our method greatly enhances the generation effect despite utilizing significantly limited trainable parameters.
These results suggest that under resource-constrained scenarios, combining minimal parameter updates with reward feedback offers a viable paradigm for achieving efficient training.

\begin{table}[htbp]
    \centering
    \small
    \setlength{\tabcolsep}{5pt}
    \resizebox{\linewidth}{!}{
    \begin{tabular}{llcccccccc}
        \toprule
        & & \multicolumn{4}{c}{\textbf{Image2Video Task}} & \multicolumn{4}{c}{\textbf{Text2Video Task}} \\
        \cmidrule(lr){3-6} \cmidrule(lr){7-10}
        \textbf{Model Size} & \textbf{Method} 
        & \textbf{CLIP ($\uparrow$)} & \textbf{LPIPS ($\downarrow$)} & \textbf{FID ($\downarrow$)} & \textbf{FVD ($\downarrow$)}
        & \textbf{CLIP ($\uparrow$)} & \textbf{LPIPS ($\downarrow$)} & \textbf{FID ($\downarrow$)} & \textbf{FVD ($\downarrow$)} \\
        \midrule
        \multirow{3}{*}{Wan2.1-1.3B} 
        & Fine-Tuning & 0.907 & 0.380 & 37.719 & 284.168 & 0.660 & \textbf{0.654} & 103.534 & 824.528 \\
        & LoRA     & 0.918 & \textbf{0.354} & 43.196 & \textbf{269.021} & 0.680 & 0.663 & 101.766 & 853.554 \\
        & Ours     & \textbf{0.923} & 0.371 & \textbf{29.236} & 274.880 & \textbf{0.684} & 0.661 & \textbf{86.560} & \textbf{637.466} \\
        \midrule
        \multirow{3}{*}{Wan2.2-5B} 
        & Fine-Tuning & \textbf{0.912} & 0.380 & \textbf{29.019} & 328.852 & 0.678 & 0.689 & 104.840 & 724.924 \\
        & LoRA     & 0.911 & 0.378 & 30.378 & 289.879 & 0.699 & 0.665 & \textbf{90.734} & \textbf{662.144} \\
        & Ours     & 0.910 & \textbf{0.366} & 29.797 & \textbf{284.780} & \textbf{0.699} & \textbf{0.663} & 94.241 & 694.937 \\
        \midrule
        \multirow{3}{*}{Wan2.1-14B} 
        & Pre-trained & 0.924 & 0.380 & 29.411 & 247.651 & 0.670 & 0.687 & 101.523 & 778.461 \\
        & LoRA     & \textbf{0.933} & \textbf{0.351} & \textbf{25.781} & 256.390 & 0.669 & 0.674 & 107.738 & 759.532 \\
        & Ours     & 0.930 & 0.356 & 28.378 & \textbf{240.456} & \textbf{0.689} & \textbf{0.668} & \textbf{94.152} & \textbf{682.851} \\
        \bottomrule
    \end{tabular}
    }
    \vspace{0.5em}
    \caption{Performance comparison of different model sizes and methods on Image2Video and Text2Video tasks. $\uparrow$ indicates higher is better, and $\downarrow$ indicates lower is better.}
    \label{tab: all_tasks}
\end{table}

\paragraph{Cross-Scale Scalability Evaluation.}
To further evaluate the scalability and generalization ability of MagicPrompt, we conduct experiments across three model sizes, namely 1.3B, 5B, and 14B parameters, on both Image-to-Video and Text-to-Video tasks. This evaluation aims to determine whether the proposed prompt-based adaptation remains effective as the base model's capacity increases. As shown in Table~\ref{tab: all_tasks}, MagicPrompt achieves consistently competitive performance across 1.3B, 5B, and 14B models on both I2V and T2V tasks. On larger backbones, especially 5B and 14B, our method improves FVD while using far fewer trainable parameters, demonstrating strong temporal quality with minimal parameter updates. In T2V, MagicPrompt shows particularly clear advantages on the 14B model, achieving the best results across all metrics, while LoRA does not consistently improve over the pre-trained baseline. The results show that MagicPrompt consistently achieves competitive performance across different model scales while requiring significantly fewer trainable parameters than conventional parameter-efficient fine-tuning methods.

\begin{figure*}[htbp]
    \centering
    \includegraphics[width=\linewidth, keepaspectratio]{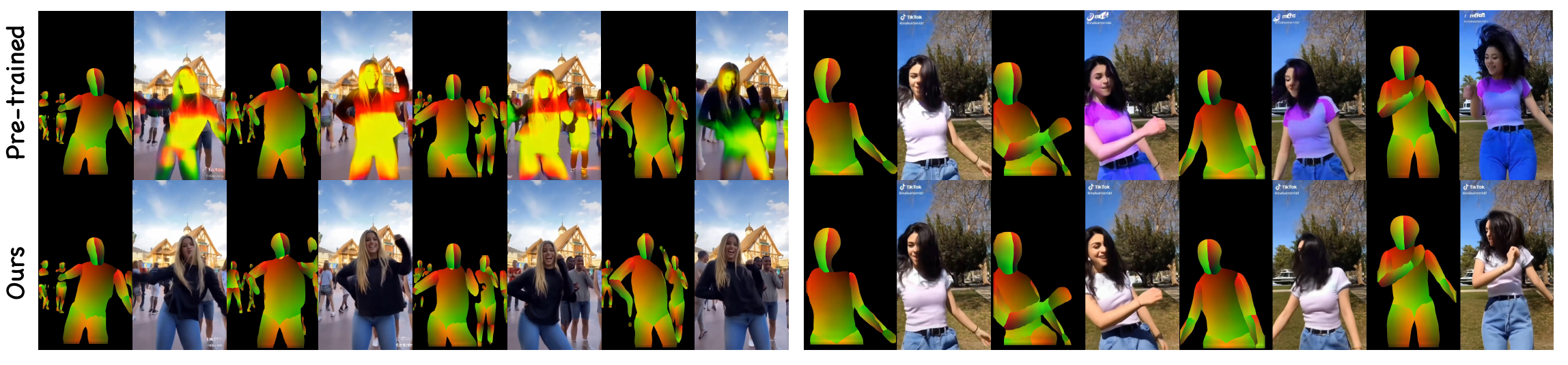}
    \caption{\textbf{New condition adaptation.} The pre-trained model suffers from notable performance degradation on unseen new conditions, as it has not learned to model the novel condition prior. After tuning with our method, the model achieves substantial performance gains and effectively adapts to the target condition.}
    \label{fig: new_condition}
\end{figure*} 

\subsection{Qualitative Experiments}

\begin{figure}[htbp]
    \centering
    \includegraphics[width=\linewidth, keepaspectratio]{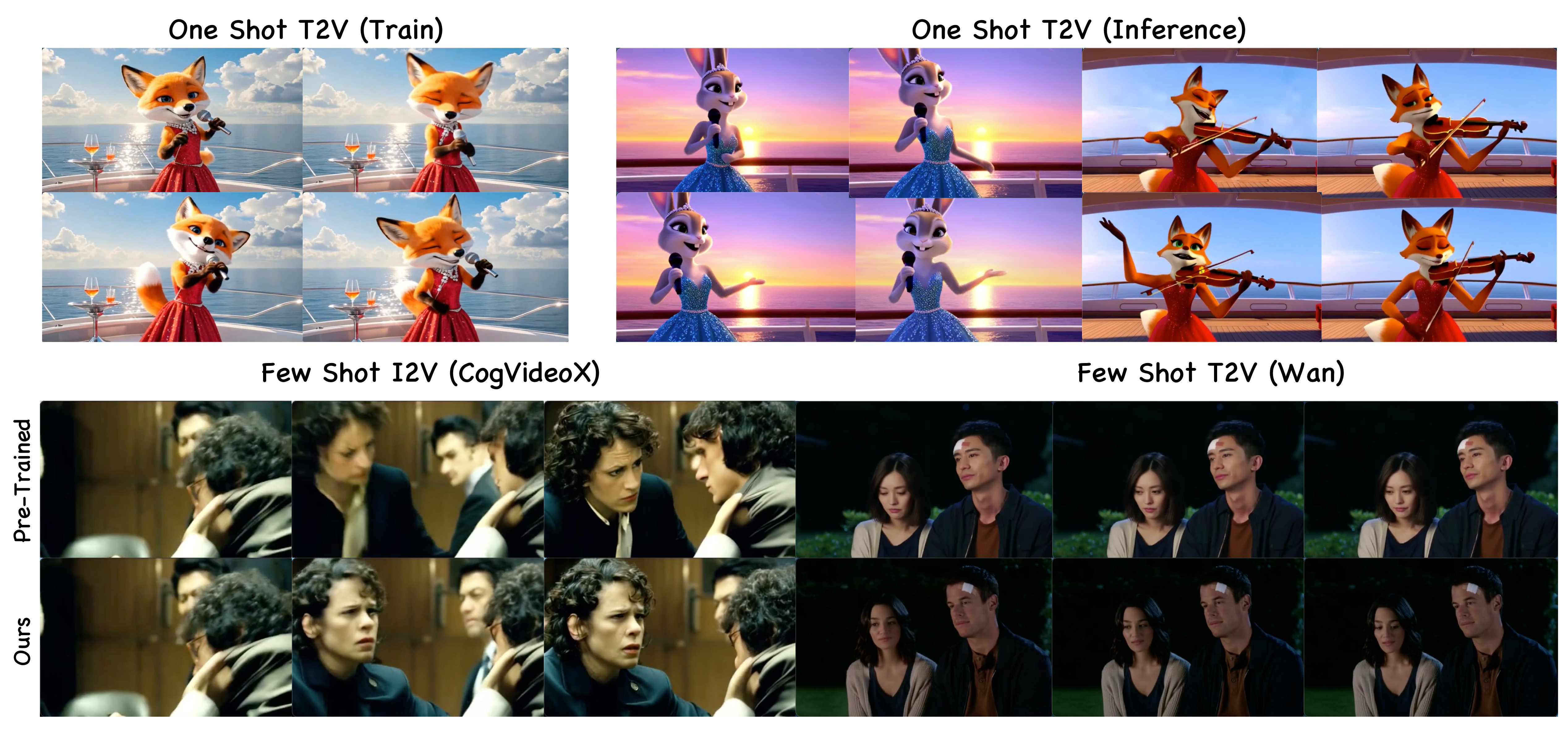}
    \caption{\textbf{Few-Shot result visualization.} Under one-shot and few-shot training regimes, our method adaptively adjusts the denoising distribution and consistently achieves better generation quality.}
    \label{fig: fewshot}
\end{figure}

\paragraph{Text-to-Video (T2V).}
As shown in Fig.~\ref{fig: visual_comparison}, our method demonstrates superior adherence to textual prompts. In the sled scenario, baseline methods such as LoRA and VACE struggle to align motion semantics with the text, leading to inconsistent character movements. In the Spiderman scenario, competing methods exhibit severe motion blur and character distortion, whereas MagicPrompt maintains sharp details and consistent identity throughout the sequence. This confirms that our attention-embedded prompts effectively guide semantic generation without disrupting spatial structure.
\paragraph{Image-to-Video (I2V).}
MagicPrompt excels in motion coherence and content preservation. In the racing car scenario, our method generates realistic, rapid motion dynamics, while baselines produce overly static or slow-moving results. In the anime scenario, MagicPrompt better preserves reference image details, such as the object held in the character's left hand, which is often lost or distorted by other methods. 
As illustrated in Fig.~\ref{fig: fewshot}, our method adaptively adjusts the denoising distribution and consistently achieves better generation quality under a few-shot scenario.
\paragraph{Control-to-Video (Control2V).}
For condition-guided generation, our method achieves robust motion consistency. As shown in Fig.~\ref{fig: new_condition} and Fig.~\ref{fig: visual_comparison}, MagicPrompt not only accurately follows the pose sequence but also adapts to new conditions without introducing structural distortion, unlike baselines that suffer from significant color blur and temporal jitter.  
These visual results corroborate our quantitative findings, demonstrating that MagicPrompt delivers high-quality, temporally coherent videos across diverse adaptation scenarios.

\subsection{Ablation Study}

To validate the design choices of MagicPrompt, we conduct ablation studies on two critical aspects: (1) the contribution of core components, and (2) the sensitivity to the number of soft tokens. All experiments are performed on the Image-to-Video (I2V) task with the Wan2.1-1.3B model.

\paragraph{Component Analysis.}
As shown in the left panel of Table~\ref{tab: ablation}, removing either core component leads to performance degradation, confirming their complementary roles. 
Specifically, \textit{w/o soft prompt} causes moderate declines in FID and FVD, indicating that prompt injection is essential for fine-grained semantic adaptation. 
More critically, \textit{w/o shift bias} results in substantial drops across all metrics, particularly FID and FVD. 
This demonstrates that the shift bias plays a crucial role in stabilizing latent distribution alignment. 
The full model, combining both components with 64 soft tokens, achieves the best overall performance, validating our design principle of non-intrusive adaptation with attention distribution regularization.

\paragraph{Soft Token Number Sensitivity.}
The right panel of Table~\ref{tab: ablation} examines the impact of soft token count on performance-efficiency trade-offs. With only 4 tokens, the model lacks sufficient capacity to capture task-specific features, resulting in suboptimal FID and FVD. 
Increasing the token number from 4 to 8 or 32 does not yield monotonic improvements, while 64 tokens provide a clear performance gain.
Notably, 64 tokens achieve the best results across all metrics, suggesting that this capacity is sufficient to encode rich adaptation signals without overfitting. 
Importantly, even with 64 tokens, our trainable parameters remain orders of magnitude fewer than LoRA or Adapter-based methods, maintaining the core advantage of parameter efficiency.

\begin{table}[tb]
  \centering
  \small
  \resizebox{\linewidth}{!}{
    \begin{tabular}{@{}lcccc@{\hspace{0.8em}}lcccc@{}}
      \toprule
      \multicolumn{5}{c}{Component Ablation} & \multicolumn{5}{c}{Soft Token Number} \\
      \cmidrule(r){1-5} \cmidrule(l){6-10} 
      Method          & CLIP $\uparrow$ & LPIPS $\downarrow$ & FID $\downarrow$ & FVD $\downarrow$ & Method & CLIP $\uparrow$ & LPIPS $\downarrow$ & FID $\downarrow$ & FVD $\downarrow$ \\
      \midrule
      w/o soft prompt & 0.91  & 0.37 & 30.94 & 284.16  & 4 token & 0.91 & 0.39 & 32.76 & 347.43 \\
      w/o shift bias  & 0.89 & 0.39 & 36.23 & 365.64 & 8 token & 0.91 & 0.40 & 34.73 & 351.18    \\
      Full(64 token) & \textbf{0.92} & \textbf{0.37} & \textbf{29.23} & \textbf{274.88}         & 32 token & 0.90 & 0.40 & 33.88 & 353.72  \\
      Fine-Tuning & 0.90 & 0.38  & 37.71  & 284.168          & 64 token    & \textbf{0.92} & \textbf{0.37} & \textbf{29.23} & \textbf{274.88}           \\
      \bottomrule
    \end{tabular}
  }
  \vspace{0.5em}
  \caption{Ablation study on components and soft token numbers.}
  \label{tab: ablation}
\end{table}
\section{Discussion}

\paragraph{Text-to-Video Task.}
When training the Wan2.1-T2V-14B model, we observe that soft prompts tend to capture and learn subject-identity information. Compared with the pre-trained baseline, the fine-tuned model frequently alters the identity of generated characters after training. Our ablation study further reveals that both soft prompts and shift bias contribute to this identity shift, and their mutual interaction amplifies this phenomenon.

\paragraph{Image-to-Video Task.}
We conduct corresponding experiments on the Wan2.2-I2V-A14B model and find that our method notably reshapes the model’s generation behavior: the model becomes more inclined to produce shot transitions, rather than generating content strictly conditioned on the initial frame. Further investigations identify shift bias as the primary source of this behavior. Removing shift bias makes the generation pattern similar to that of the pre-trained model, while retaining shift bias enables the model to produce multi-shot variations based on the reference information from the initial frame, including changes in characters, scenery, and other scene elements. The specific manifestation of this effect also varies slightly across different training datasets. In addition, we perform ablation studies on the injection position of soft prompts, comparing prefix injection and suffix injection. The two strategies yield similar overall effects, while prefix injection achieves marginally better performance.

\section{Conclusion}
\label{sec:conclusion}

We introduced MagicPrompt, an ultra-lightweight framework for adapting large-parameter video diffusion models with minimal trainable parameters ($<1\%$).
Experiments across diverse video tasks demonstrate that MagicPrompt achieves competitive visual results, compared to existing parameter-efficient methods, while reducing trainable parameters by orders of magnitude.

{
    \small
    \bibliographystyle{ieeenat_fullname}
    \bibliography{main}
}

\end{document}